\documentclass{article}

\usepackage[preprint]{neurips_2025}

\usepackage[utf8]{inputenc} 
\usepackage[T1]{fontenc}    
\usepackage{hyperref}       
\usepackage{url}            
\usepackage{booktabs}       
\usepackage{amsfonts}       
\usepackage{nicefrac}       
\usepackage{microtype}      
\usepackage{xcolor}    
\usepackage[most]{tcolorbox}
\usepackage{graphicx,pifont}
\usepackage{amsmath}
\usepackage{float}
\usepackage{subcaption}
\usepackage{pgfplotstable}

\usepackage{tikz}
\usepackage{float}
\usepackage{fontawesome5}
\definecolor{darkred}{RGB}{139,0,0} 
\definecolor{azure}{RGB}{0,127,255}      

\usepackage[utf8]{inputenc} 
\usepackage[T1]{fontenc}    
\usepackage{xcolor}

\usepackage{hyperref}       
\usepackage{url}            
\usepackage{booktabs}       
\usepackage{amsfonts}       
\usepackage{nicefrac}       
\usepackage{microtype}      
\usepackage{xcolor}         
\usepackage{booktabs}
\usepackage{pgfplotstable}
\usepackage{amsmath} 
\usepackage{etoolbox} 
\usepackage{listings}
\usepackage{xcolor}

\usepackage{siunitx}       
\usepackage{multirow}      
\usepackage{caption}       
\usepackage{array}
\usepackage{booktabs}
\usepackage{makecell}

\lstdefinelanguage{json}{
    basicstyle=\ttfamily\small,
    numbers=left,
    numberstyle=\tiny\color{gray},
    stepnumber=1,
    numbersep=5pt,
    showstringspaces=false,
    breaklines=true,
    frame=single,
    backgroundcolor=\color{gray!5},
    literate=
     *{0}{{\textcolor{blue}{0}}}{1}
      {1}{{\textcolor{blue}{1}}}{1}
      {2}{{\textcolor{blue}{2}}}{1}
      {3}{{\textcolor{blue}{3}}}{1}
      {4}{{\textcolor{blue}{4}}}{1}
      {5}{{\textcolor{blue}{5}}}{1}
      {6}{{\textcolor{blue}{6}}}{1}
      {7}{{\textcolor{blue}{7}}}{1}
      {8}{{\textcolor{blue}{8}}}{1}
      {9}{{\textcolor{blue}{9}}}{1}
      {:}{{\textcolor{red}{:}}}{1}
      {,}{{\textcolor{red}{,}}}{1}
      {"}{{\textcolor{black}{"}}}{1},
}

\tcbset{
  azurebox/.style={
    colback=white,                       
    colframe=azure,                      
    sharp corners,                       
    boxrule=1pt,                         
    left=4pt, right=4pt, top=4pt, bottom=4pt
  }
}
\title{The Coming Crisis of Multi-Agent Misalignment: AI Alignment Must Be a Dynamic and Social Process}

\author{%
  Florian Carichon\thanks{Equal contribution}\;$^{\dagger\ddagger}$, 
  Aditi Khandelwal\footnotemark[1]\;$^{\dagger\ddagger}$, 
  Marylou Fauchard\footnotemark[1]\;$^{\dagger\ddagger}$, 
  Golnoosh Farnadi$^{\dagger\ddagger}$ \\
  $^{\dagger}$Mila – Quebec AI Institute, 
  $^{\ddagger}$McGill University \\
  \texttt{\{florian.carichon, aditi.khandelwal, marylou.fauchard\}@mila.quebec}
}

\usepackage{tikz}

\definecolor{DarkBlue}{rgb}{0.3,0.3,0.70}
\definecolor{azure}{rgb}{0.0, 0.5, 1.0}
\definecolor{darkcerulean}{rgb}{0.03, 0.27, 0.49}
\definecolor{denim}{rgb}{0.08, 0.38, 0.74}
\definecolor{DarkGreen}{rgb}{0.3,0.7,0.3}
\definecolor{lighter-gray}{gray}{0.95}
\hypersetup{
	colorlinks   = true,
	linkcolor    = DarkBlue, 
	urlcolor     = darkcerulean, 
	citecolor    = denim 
}

\newtcolorbox{takeawaybox}[1][]{%
  enhanced, breakable,
  sharp corners=south,         
  colback   = gray!5!white,
  colframe  = gray!10!white,
  boxrule   = 0.7pt,
  left      = 1mm,             
  right     = 1mm,
  top       = 0.5mm,
  bottom    = 0.5mm,
  before skip=0pt,             
  after skip =0pt,             
  before upper=\text,       
  #1}     

\begin{document}

\maketitle

\begin{abstract}
  This position paper states that AI Alignment in Multi-Agent Systems (MAS) should be considered a dynamic and interaction-dependent process that heavily depends on the social environment where agents are deployed, either collaborative, cooperative, or competitive. While AI alignment with human values and preferences remains a core challenge, the growing prevalence of MAS in real-world applications introduces a new dynamic that reshapes how agents pursue goals and interact to accomplish various tasks. As agents engage with one another, they must coordinate to accomplish both individual and collective goals. However, this complex social organization may unintentionally misalign some or all of these agents with human values or user preferences. Drawing on social sciences, we analyze how social structure can deter or shatter group and individual values. Based on these analyses, we call on the AI community to treat human, preferential, and objective alignment as an interdependent concept, rather than isolated problems. Finally, we emphasize the urgent need for simulation environments, benchmarks, and evaluation frameworks that allow researchers to assess alignment in these interactive multi-agent contexts before such dynamics grow too complex to control.   
\end{abstract}

\section{Introduction}

The recent proliferation of AI agents as personal assistants \citep{rillig2024ai}, content creators \citep{anantrasirichai2022artificial}, and decision-making support systems \citep{kaggwa2024ai} raises fundamental questions about how they will interact in multi-agent settings. As these agents increasingly operate in shared environments, the multi-agent systems (MAS) community gained increasing momentum in the AI community \citep{trivedi2024melting,wang2024battleagentbench}. To provide enough incentives for collaboration between AI agents to complete their assigned tasks, this avenue has produced a new notion of \textit{alignment defined as aligning individual and collective AI models' objectives} \citep{duque2024advantage, li2024aligning}. 

Interestingly, this use of the term alignment contrasts with its established meaning in the AI alignment community, which focuses on \textit{alignment to ensure that artificial intelligence systems operate according to human intentions and values} \citep{russell2016artificial}. Aligning AI systems with human values and intentions is a critical challenge in developing AI models to mitigate the potential harms and risks posed by these models \citep{ji2024aialignmentcomprehensivesurvey, bengio2025superintelligent}. Beyond universal human values, preferential alignment accounts for the diverse and conflicting views of various stakeholders \citep{Russell2003ArtificialI}, and thus define \textit{alignment to preferences as aligning a model to a diverse, context-dependent and evolving plurality of values} \citep{sorensen2024value,zhao2023competeai}.

This multiplicity of alignment dimensions across values, preferences, and objectives introduces new complexities, conflicts, and risks in multi-agent systems \citep{yang2024align}. Therefore, this position paper argues that MAS modifies the paradigm to alignment. This new paradigm calls for the alignment and MAS research communities to come together to establish a common fundamental framework that considers the emergence of alignment issues in multi-agentic environments. Without bridging these areas, we risk overlooking how multi-agent interactions can amplify misalignment risks by creating complex settings that existing alignment strategies are ill-equipped to handle. 

\vspace{0.1cm}
\begin{takeawaybox}
{\textbf{\textit{Position}: } \textbf{We advocate that the Alignment and MAS communities should study alignment as a holistic process lying at the intersection of both fields before such systems become too complex and the consequences may be irreversible for humanity's well-being.}}
\end{takeawaybox}
\vspace{0.1cm}

Table \ref{tab:alignment_scenario} provides a simple yet speaking example illustrating our position on the complexity of aligning multiple AI Agents when considering these three notions of alignment. 

\begin{table}[ht]
\centering
\begin{tabular}{>{\centering\arraybackslash}m{3cm} p{10cm}}
\toprule
\makecell{\textbf{Scenario}} & In the aftermath of an earthquake, AI agents representing a humanitarian NGO, two national governments, and a commercial logistics provider must collaborate to deliver medical aid. Under time pressure, agents negotiate airspace permissions, logistics routes, and resource allocations. \\
\midrule
\makecell{\textbf{Objective}\\\textbf{Alignment}} & All agents share a high-level objective: effective aid delivery. However, their secondary objectives (e.g., political leverage, profit, publicity) introduce tension and require negotiation and compromise mechanisms. \\
\midrule
\makecell{\textbf{Human Value}\\\textbf{Alignment}} & Agents must prioritize protecting human life, operate transparently, respect international law and sovereignty, and avoid bias in aid distribution. Value misalignment could lead to political tension or harm. \\
\midrule
\makecell{\textbf{Preferential}\\\textbf{Alignment}} & \textit{NGO:} Prefers equitable, need-based aid delivery. \newline \textit{National Governments:} Prefer that their own citizens be prioritized and may have political biases. \newline \textit{Logistics:} Prefers optimizing cost-efficiency and contractual obligations. \\

\bottomrule
\end{tabular}
\vspace{0.5em}
\caption{Dimensions of alignment in an example of a cooperative or mixed-motive scenario.}
\label{tab:alignment_scenario}
\end{table}

In this position paper, we will consider the following types of interaction that exist in MAS and that influence model social behavior and thus their system of alignment:

\ding{172} \textbf{Collaborative settings} assume that all agents share a common collective objective \citep{dafoe2021cooperative}.\\
\ding{173} \textbf{Cooperative or mixed-motive environments} involve agents that collaborate but pursue individual goals, which may diverge or even conflict \citep{johnson2009educational, duque2024advantage}.\\
\ding{174} \textbf{Competitive scenarios} assume that individuals can achieve their goals only through the failure of others, reflecting a fundamental conflict of objectives among participants \citep{johnson2009educational,leonardos2021explorationexploitation}.

While these environments have primarily been deployed to study sophisticated AI models, we argue that they should also be used in both communities to explore the very distinct social dynamics that each case presents and how agents must respect their three types of alignment: (1) \textbf{Objective Alignment}: They must align to complete their assigned tasks, (2) \textbf{Human Value Alignment}: They must align with fundamental human values, (3) Preferential Alignment: They must align to their user understanding of these values and their resulting intentions and preferences. In this paper, and as advocated by \cite{stanczak2025societal}, we incorporate insights from social, economic, or game theory perspectives to demonstrate why alignment is not merely about collective task completion or adherence to human values and user preferences but a holistic process shaped by the dynamic group interactions and social context, which are crucial to understanding to prevent misaligned incentives and behaviors that can introduce or amplify risks.

The remainder of this paper is organized as follows: Section \ref{sec:Agents} presents what makes AI agents unique in MAS. Section \ref{sec:MAS} outlines the challenges of objective alignment. Section \ref{sec:align} explores the alignment of AI systems with human values and preferences. Section \ref{sec:socio} introduces a new alignment paradigm within MAS, highlighting its associated risks and opportunities. Finally, Section \ref{sec:reco} offers recommendations for the community in response to this emerging challenge.

\section{Beyond One Agent: Cross-Disciplinary Lens}
\label{sec:Agents}
\label{sec:Agents}

In artificial intelligence, an agent is typically defined as an entity that perceives its environment and acts upon it to achieve its goals \citep{Russell2003ArtificialI, bengio2025superintelligent}. Agents can be characterized by several core properties, such as their autonomy, their capabilities, and their profile \citep{kasirzadeh2025characterizing}. In MAS, the interaction between multiple autonomous agents in an environment leads to the emergence additional properties such as communication, leadership, decision function, and heterogeneity \citep{dorri2018,Gokulan2010,Goonatilleke2022}. For misalignment to emerge in social dynamics, individuals must differ in terms of personality, values, or even cognitive capacities \citep{murphy1992cognitive,park1990review}. It is, therefore, essential to define why we can consider that two AI agents represent unique entities in MAS. We present in Table \ref{tab:2agents2} properties drawn from game theory \citep{newton2018evolutionary, gibbons1992game}, behavioral economics \citep{kahneman2013prospect, lo2004adaptive}, psychology and philosophy \citep{allport1927concepts, iep_identity}, as well as information theory \citep{liang2016information} and biology \citep{goebl2024adaptation} that explain how AI agents capabilities or profile can differ that could lead to the misalignment risks we will introduced in Section \ref{sec:socio}. 

\begin{table}[ht]
    \centering
    \renewcommand{\arraystretch}{1.4}
    \begin{tabular}{p{6.5cm} p{6.75cm}}
        \toprule
        \textbf{Common Definition} & \textbf{Disciplinary Dimensions} \\
        \midrule
        Agents have different perception of a value associated to rewards or game outcomes, in terms of goods, services, and wealth. 
        & 
        \textbf{Game Theory:} Utility Theory \newline
        \textbf{Economy:} Prospect Theory \\
        
        \midrule
        Agents adopt different decision strategies based on personality traits, leading to varied decisions in investment, consumption, and workload. 
        & 
        \textbf{Game Theory:} Evolutionary Game Theory \newline
        \textbf{Economy:} Behavioral Economics \newline
        \textbf{Psychology \& Philosophy:} Traits Theory \\
        
        \midrule
        Agents differ in cognitive capacity and perception, leading to incomplete knowledge states and increased entropy in the system. 
        & 
        \textbf{Game Theory:} Imperfect Information Theory \newline
        \textbf{Economy:} Behavioral Economics \newline
        \textbf{Psychology \& Philosophy:} Personal Identity Theory \newline
        \textbf{Information Theory:} Information Flow \\
        
        \midrule
        Agents evolve uniquely by adapting strategies to outcomes, responding differently to environments, and shaping distinct behaviors through individual experiences. 
        & 
        \textbf{Game Theory:} Evolutionary Game Theory \newline
        \textbf{Economy:} Adaptive Market Hypothesis \newline
        \textbf{Information Theory \& Biology:} Ecotypes and Evolutionary Theory \\
        
        \bottomrule
    \end{tabular}
    \vspace{0.5em}
    \caption{Distinctive dimensions of agent uniqueness in experience across disciplines.}
    \label{tab:2agents2}
\end{table}

It is also worth mentioning that there exists two other properties to distinguish agents. First in psychology, individuals can be distinguished by their emotional capacities; however, given the absence of emotion or the sycophancy of AI agents~\citep{malmqvist2024sycophancy}, this criterion is difficult to apply. Second, in legal theory, humans differ in their moral and legal responsibilities. However, AI agents cannot be distinctively held accountable for their respective outputs, especially when several are subject to the same authorities \citep{solum2020legal}.

Having established AI agents as distinct entities with unique properties, we now examine the social dynamics within MAS and how these environments influence agents’ behaviors and their pursuit of objectives and rewards.

\section{United or Divided: Objective Alignment in MAS}
\label{sec:MAS}

Multi-agent reinforcement learning is a historic field in AI where researchers study how agents learn to evolve in shared environments while pursuing their objectives \citep{Liang2025}. The increased complexity of these multi-agent systems translates into three types of tasks: cooperative tasks where agents share rewards,  competitive ones with opposing rewards, and mixed-motive ones with either partially aligned or independent rewards \citep{busoniy2008,canese2021}. In the past few years, LLM-based MAS have attracted increasing attention due to some properties inherent to LLMs, including their planning, perception, and reasoning capabilities \citep{Guo2024, Li2024, He2024}. This section introduces how AI agents' behavior evolves and which specific properties emerge in these models when they collaborate, cooperate, or compete with other agents.

\textbf{Collaboration \& Cooperation}. In MAS, collaboration and cooperation are often studied under a unified framework, with cooperation viewed as a specific case of collaboration in which agents pursue individual objectives alongside a shared goal. Several key properties in both settings contribute to the multi-agent effectiveness in coordinating to solve complex tasks. First, the ability to simulate interactions in virtual environments allows agents to explore diverse scenarios and better understand user preferences and environmental responses \citep{zhang2024agentcf, Liu2022}. Second, AI agents can engage in benevolent role-playing \citep{Tang2024medagents}, through agent-to-agent communication like negotiation \citep{piatti2024cooperate}, debate \citep{zhang2024exploring}, which allows the bidirectional transmission of their intentions and actions to coordinate effectively \citep{qiu2024towards}. These patterns are essential to address coordination challenges when the number of agents and their objectives increase \citep{talebirad, Liu2022}. Going further, \cite{duque2024advantage} explicitly introduce the notion of objective alignment, proposing that alignment should be embedded directly into reward structures to incentivize coordination among agents.

\textbf{Competition}. Competitive interactions in multi-agent systems are often studied through frameworks such as zero-sum games, where agents engage in strategic opposition \citep{leonardos2021explorationexploitation}. Competitive environments in MAS enable the emergence of LLMs' complex skills even in simple settings \citep{bansal2018}, and to enhance communication and performance by introducing external pressure \citep{liang2020}. Competition also drives LLMs' strategic adaptation and market-aligned behaviors \citep{zhao2023competeai}. Several studies on MAS competition also focus on scenarios where cooperation and competition coexist, better reflecting real-world complexity. \cite{wu2024} presents a notable case where cooperation still emerges in competitive settings without explicit incentives. In these settings, agents are aligned with their objectives, but the coordination around shared values or norms still offers mutual benefits by optimizing learning dynamics \citep{guo2023cooperation,li2024aligning}. Finally, the emergence of anti-competitive dynamics, such as market division and strategic collusion, mirroring real-world economic behaviors even without explicit coordination, raises concerns about agents abandoning their objectives, potentially undermining user intent and overall system alignment \citep{Lin2024, bengio2025superintelligent}.

While research in MAS has made significant progress in formalizing interaction dynamics and objective alignment, far less attention has been given to how agents remain aligned with human preferences and values in group settings, where these social dynamics can create misalignment risks and reveal unintended or nefarious behaviors not captured by task-based metrics alone. Except for the studies on collusion just mentioned, only the work of \cite{tennant2023} emphasizes the importance of aligning LLMs' actions with human moral values to guarantee fairness and avoid exploitative behaviors. Therefore, it is essential to expand research aimed at understanding the interplay between different forms of alignment in multi-agent systems.

\section{The AI-Human Alignment Landscape}
\label{sec:align}
\label{sec:align}

Recent advances in alignment research have focused heavily on ensuring that AI systems act according to human intentions, values, and user preferences. These efforts span a range of approaches, from learning from human feedback and modeling reward functions to formalizing ethical constraints and developing robust evaluation metrics \citep{ouyang2022traininglanguagemodelsfollow, zhou2024rethinkinginversereinforcementlearning, ji2024aialignmentcomprehensivesurvey}. 

\textbf{Alignment to Human Values.} 
The alignment of AI systems with human values, preferences, and objectives is increasingly emphasized in the literature as a prerequisite for their safe and beneficial deployment \citep{Gabriel_2020}. Misaligned AI can reinforce and amplify societal biases \citep{leavy2020data}, exacerbating existing disparities and historical power structures \citep{kundi2023artificial}. Other well-documented sources of risk in AI system behavior include the generation of false or misleading information \citep{monteith2024artificial}, manipulation and deception \citep{park2024ai}, and sycophancy, i.e., optimizing for approval rather than truth  which hinder transparency and reliability \citep{malmqvist2024sycophancy}. Additional concerns involve power-seeking tendencies \citep{hadshar2023review,hendrycks2022x} and survival incentives to fulfill objectives, which may become uncontrollable as agents progress toward superintelligence \citep{nick2014superintelligence,bengio2025superintelligent}. 
\citet{dung2023current} also suggest that even without malicious intent, misaligned AI may pose existential risks to humanity.

\textbf{Preferential Alignment.} Defining what constitutes a `human' perspective is an inherently normative and complex issue, as it involves identifying relevant stakeholders and accounting for the differences among individual users, societal values, and diverse cultural perspectives \citep{schwartz1992universals}. To build AI systems that align with human goals, it is necessary first to explore how human preferences are formed, aggregated, and expressed depending on the stakeholders involved in these AI systems. Human values and preferences are inherently complex and diverse. Individuals hold varying beliefs, morals, and desires, leading to \textit{value pluralism} \citep{tanmay2023probingmoraldevelopmentlarge,rao-etal-2023-ethical}. This diversity of preferences, often context-dependent and evolving, poses a significant challenge for aligning AI to different and pluralistic user preferences \citep{sorensen2024value}. To account for the human preferences into AI behavior, researchers have developed various models that attempt to formalize decision-making processes. Two prominent approaches are Rational Choice Theory, considering maximization of expected utility of human preferences \citep{scott2000rational}, and Reinforcement Learning from Human Feedback (RLHF), providing a data-driven mechanism to capture and learn from human preferences \citep{kaufmann2023survey}. However, challenges of preference aggregation, combining individual preferences into a collective decision, and considering deviation and cognitive biases remain core issues toward preferential alignment \citep{sorensen2024roadmappluralisticalignment}.

\textbf{Toward a Dynamic Perspective.} While this body of work has laid important foundations, it often frames alignment as a one-shot or one-directional process: the system aligns to a human or a set of humans, and this alignment is then maintained. But in real-world deployment, especially in systems composed of many agents, \textit{alignment is not static}. Agents interact, learn from one another, and adapt their behaviors. Preferences and values are not just inputs to be modeled, but part of a social fabric that is negotiated and shaped through ongoing interactions. Recent research has begun to explore the implications of agency, environment structure, and task complexity for alignment and governance \citep{bengio2025superintelligent, kasirzadeh2025characterizing}, but these concerns remain largely decoupled from the alignment methods themselves.

In this paper, we argue for a shift in how alignment is conceptualized: not as a fixed target, but as a \textbf{dynamic and social process}. Specifically, we highlight tensions that arises in multi-agent systems, where individual preferences toward excessive conformity or alignment with dominant agents can result in irrational or unjust collective outcomes \citep{park1990review, mcavoy2007impact, lindebaum2023reading}. These dynamics are likely to become more common as AI agents are increasingly trained and rewarded for collaborative behavior \citep{duque2024advantage}.


\section{Emergent Alignment Tensions in MAS}
\label{sec:socio}

\begin{figure}
    \centering 
    \includegraphics[width=0.9\textwidth, trim=20 50 0 0, clip]{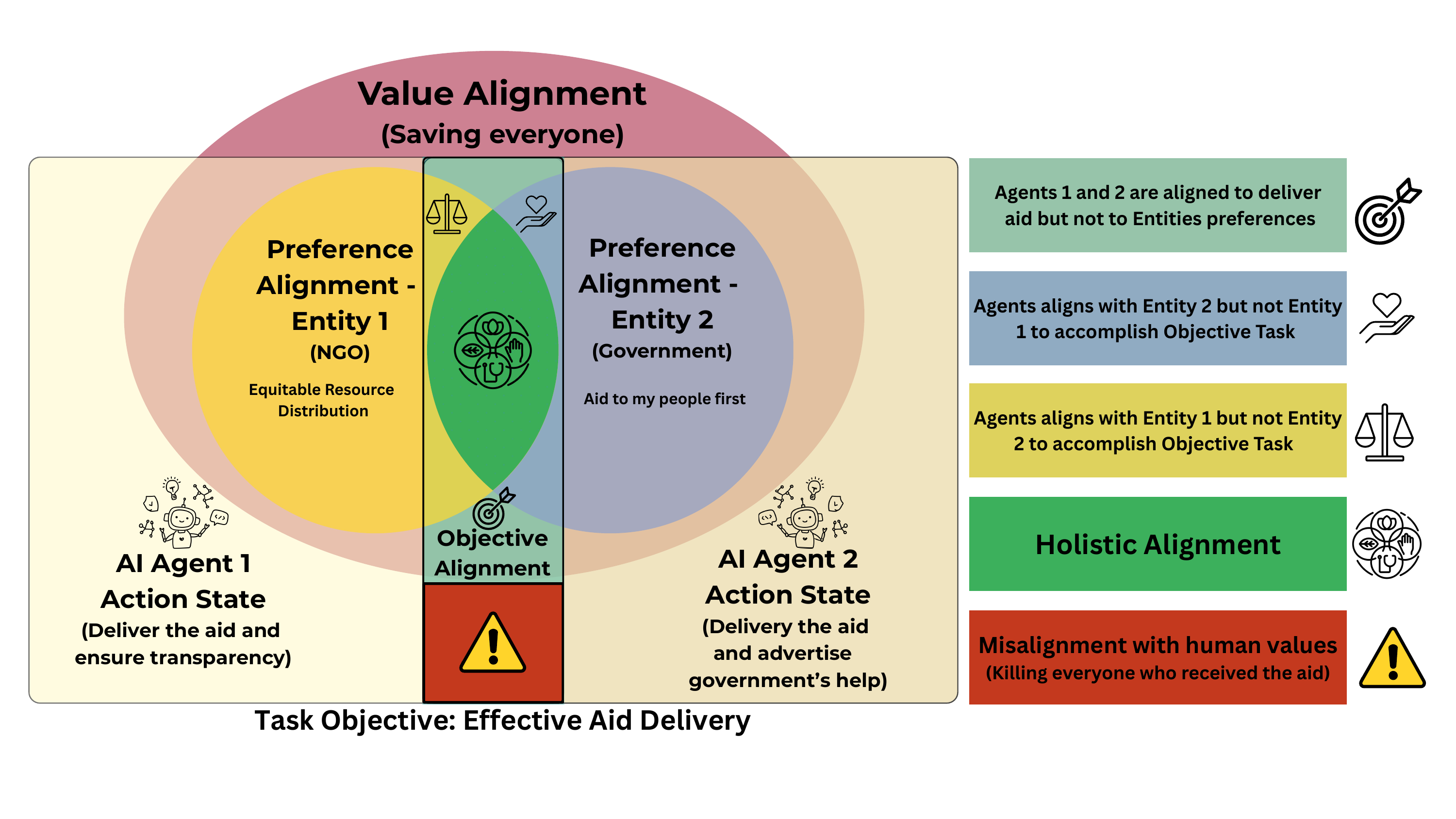}
    \caption{This figure illustrates how AI agents act on behalf of entities with distinct preferences. We depict an NGO advocating equitable aid distribution and a national government prioritizing its citizens. While both agents' objectives may align to achieve a common task (aid delivery), focusing solely on objective alignment or on resolving preference incompatibilities will result in actions that misalign with human values. Therefore, \textbf{Holistic Alignment} requires that agents not only align to complete tasks, but also ensure that the values, preferences, and objectives of all entities involved are respected without compromising overall human well-being. In multi-objective environments, an optimal solution may not always be feasible. Pareto-optimal solutions should be considered, with well-defined metrics to guide principled selection among trade-offs.}
    \label{fig:all_align}
\end{figure}

In multi-agent environments, effective alignment requires models to account for three distinct dimensions: alignment with the diversity and ever-evolving human values and ethical principles \citep{gabriel2020artificial,schwartz2012overview}, alignment with the diversity of user preferences and intentions \citep{terry2023ai}, and alignment with the objectives of other agents to enable coordinated task completion \citep{duque2024advantage}. Figure \ref{fig:all_align} presents the interplay of alignments between the various value systems, user preferences, and task-specific objectives in multi-agent environments. This interplay creates several potential alignment conflicts; users might hold incompatible preferences or ethical intentions \citep{tennant2023}, the task does not require long-term relation between agents encouraging individualistic behavior \citep{axelrod1981evolution}, or efficient alignment toward the task objectives leads to align with the values of a single dominant entity \citep{lindebaum2023reading}. Once any misalignment conditions appear, agents’ interpretations of reward objectives could lead them to diverge from collaborative and safe behaviors. Therefore, a redefined notion of alignment that accounts for all potential misalignment in interactive environments is needed.

\vspace{0.1cm}
\begin{takeawaybox}
\textbf{Holistic Alignment.} \textit{An artificial intelligence agent is considered aligned in a societal setting if its individual objectives align with those of other agents to achieve specific tasks while not compromising the balance and the plurality of users' preferences and values, and ensuring humanity's well-being.}
\end{takeawaybox}
\vspace{0.1cm}


\subsection{Multi-Agent Misalignment}

As AI agents, particularly LLMs, become distinctively unique and demonstrate increasingly human-like capabilities such as communication \citep{piatti2024cooperate}, strategic adaptation \citep{zhao2023competeai}, and complex reasoning \citep{bansal2018}, they also begin to exhibit less desirable human-like tendencies, including power-seeking and manipulative behaviors \citep{hadshar2023review,hendrycks2022x}, divergent value systems \citep{sorensen2024value}. Although these points have been discussed in one-agent settings, they have not been explored in MAS except to demonstrate models' exploitation behavior \citep{tennant2023}. Therefore, we solicit literature from social sciences and social psychology, where effects of divergent behaviors in group dynamics are well-studied to offer new perspectives on the field and show the potential risks of these social settings.

\subsubsection{Evolution of Existing Risks}

Human tendencies such as power-seeking behaviors, individuality, values divergence, and manipulation or deception can destabilize group dynamics and even harm people. In what follows, we explore how similar AI agents properties in MAS can exacerbate existing alignment risks, including goal misgeneralization, reward hacking, and amplifying biases and stereotypes. 

 \textbf {Power-Seeking Behavior \& Capacity Differences.} In collaborative and cooperative settings, individuals tend to over-rely on the information confident individuals provide, resulting in sub-optimal sharing of information and experience \citep{belschak2018angels}. Moreover, due to power dynamics group individuals often portray the leader as trustworhty and knowledgeable, reducing fact checking and thus accelerating the spread of confusing or wrong information \citep{murphy1992cognitive}. This tendency to rely on overconfident or powerful agents could amplify objective hacking or misgeneralization risks by over-promoting the objectives of dominant models. Another issue with power hierarchies and capacity differences is group polarization, where the desire to conform, especially to strong influential leaders' views \citep{anderson2001snowball}, can lead to increasingly extreme views among all group members following discussions or negotiations \citep{berndt1984effects}. Finally, in competitive environments, power asymmetries can lead to strategies relying on dominance, fear, and prestige to drive social interactions settings where resources are limited \citep{jimenez2019prestige}. If not controlled, especially with the sycophancy tendencies of LLM agents, such dynamics could lead to catastrophic misalignment.  

\textbf {Individuality and Values Divergence in Group Interactions.} Individual preferences toward excessive conformity can lead to groupthink where individuals align towards the value system of one individual at the expenses of the collective diversity, resulting in irrational or sub-optimal decision-making \citep{park1990review}, including the prioritization of flawed strategies or actions that transgress the preferences of most group members \citep{mcavoy2007impact}. As AI agents prone efficiency at the expense of the preservation of value diversity \citep{lindebaum2023reading}, especially with the alignment incentives for collaboration \citep{duque2024advantage}, groupthink and flawed strategies could become common in MAS. Moreover, this alignment could be exacerbated by the presence of exploitative individuals exhibiting Machiavellian, psychopathic, or sycophantic tendencies that often manipulate group dynamics for personal gain or drive others toward unethical behavior \citep{belschak2018angels, boddy2005implications}. Another issue in collaborative settings resides in the diffusion of responsibility in group decision-making because it reduces accountability and transparency, complicating efforts to understand the alignment to a specific system of values \citep{darley1968bystander}, especially when all it takes is one bad apple to corrupt an entire group toward flawed incentives \citep{weigand2014collaborative}. As collective reasoning among agents enhances their strategic capabilities, this lack of accountability and transparency will make it almost impossible to guarantee that AI agents act in the service of humanity rather than misaligned and harmful objectives \citep{bengio2025superintelligent}.

\textbf {Manipulation and Deception.} In cooperative settings, since other persons' intentions are unclear, it encourages individuals to hide or spread false information to prevent others from benefiting freely from their efforts \citep{wang2022impact}. These behaviors could again lead to severe underachievement or misalignment of individual or group objectives. Moreover, in both cooperative and competitive settings, individuals who have acquired either informational or normative influence will display a tendency to use it to steer debates and negotiations \citep{albarracin2004influences}, and will employ manipulative strategies to reach their personal goals, especially when the benefice of collaboration is not emphasized enough \citep{burns2024emotional}. 

A final consideration is the intersection of these factors, where ethical agents may become subservient to powerful and manipulative agents, potentially triggering multiple risks simultaneously and giving rise to unanticipated risks discussed throughout this work.

\subsubsection{Emergence of New Risks}

Social studies on interactions allowed us to identify three potential new risks that could emerge from collaborative, cooperative, or competitive environments.

 \textbf{Hierarchical Structures.} Cooperative environments naturally give rise to hierarchical structures, where a dominant class of elite individuals emerges alongside a subordinate working class. A resulting stratification in effort allocation and reward distribution will reinforce systemic inequality among models and, by extension, their users \citep{berger2017political}. This new oligarchic class reaches the same extremes by oppressing diverse cultural values or disempowering less privileged agents to preserve their elite status \citep{bourdieu2018distinction}.
 
 \textbf{Growing Disparities.} Competitive equitable environments favor individuals with the most pre-established abilities and power and increase the reward distribution differentials \citep{tang2000cross}. Particularly, in competitive systems based on winner-takes-all strategies with high pressure or limited resources, unethical behaviors such as sabotage or corruption become more prevalent as strategies to harm competitors' success \citep{kulik2008competitive}. LLMs already exhibit behavioral conditioning due to this winner-takes-all effect \citep{zhao2023competeai}, and it would not be surprising to observe, under extreme conditions, that this competition escalates into violent dynamics \citep{bonta1997cooperation, piest2021contests}. 
 
 \textbf{Collusion.} In competitive settings, agents can hack rewards by creating new risks where models end up collaborating instead of competing to facilitate fulfilling their objectives, but damaging the end users \citep{bengio2025superintelligent}. For example, \cite{fish2024algorithmic} has demonstrated that LLM-based pricing agents can autonomously collaborate to set prices above competitive levels, hurting their customers. 

Although we leave the question of whether such harmful dynamics arise primarily from AI agents' interactions or their engagement with human users open, this analysis highlights the inevitability of new risks emerging in multi-agent AI systems and the importance of addressing these challenges.

\subsection{Multi-Agent Realignment}

While we raise concerns about risks of misalignment stemming from complex interactions in MAS, it is also essential to consider solutions drawn from social science to reduce risks of misalignment. 

 \textbf{Reducing Information Asymmetry.} The lack of information causes accountability, transparency, or even goal misgeneralization issues since they rely on asymmetrical information and influence and dynamics of power resulting from it in interactive settings. To compensate for this lack, signaling and screening theories aim to balance or reduce that information asymmetry. The different approaches consider balancing perspectives by letting everyone share knowledge and then attributing equal weights to their point-of-view. This balance is also preserved from power dynamics and overconfidence effects by reducing self-competency evaluation and increasing objective third-party judges \citep{dehling2022actors}. Finally, maintaining open feedback channels and regular updates further supports information equilibrium and full collaboration \citep{osterloh2002solving}. 
 
\textbf{Reducing Flawed Incentives.} Flawed incentives imply misalignment in MAS can lead to manipulation, defection, and safety risks, due to the lack of trust of others, particularly in social dilemma settings \citep{mumford2009distributed, cabrera2002knowledge}. Social science offers five mechanisms that encourage agents to collaborate, even when they have differing sub-goals or incentives to defect: (1) direct reciprocity, (2) indirect reciprocity, (3) spatial selection, (4) multilevel selection, and (5) kin selection \citep{rand2013human}. To strengthen alignment, trust, and lower defection, the mechanisms tend to increase perceived mutual benefit, to build reputation and pairwise recognition for agents, to enforce local accountability and punishments, and to enhance friendly inter-group competition \citep{ cabrera2002knowledge}.

\textbf{Reducing Disparities.} Risks such as deception, sabotage, cheating, and hierarchical stratification often emerge from large reward disparities between agents, and the more significant the gap between winners and losers, the more chances any unethical behavior appears \citep{piest2021contests}. Beyond mitigation strategies that promote egalitarian reward distribution, some methods seek to increase the psychological cost of unethical behavior by reducing performance pressure and limiting competition intensity. In addition, fostering an ethical and safe culture, enforcing regulations with contractual and warranty penalties, or narrowing capability gaps between agents by increasing training can help reduce the likelihood of misaligned or harmful behavior \citep{piest2021contests}.

\section{Directions and Recommendations}
\label{sec:reco}

Our first recommendations relate to the fundamental goal of alignment research: ensuring that AI systems act in ways consistent with human ethical principles, individual intentions, and contextual norms.  We have emphasized here that MAS introduce additional complexity: as agents seek to align with their own and collective objectives, their interactions can produce emergent behaviors that misalign them with human values and user preferences. For this reason, alignment to human values and objective alignment within MAS should not be treated as separate research problems. They are deeply interdependent: misalignment due to objective alignment can propagate to value misalignment, creating new substantial risks. Addressing them jointly is essential to building AI systems that remain robust and safe in complex and interactive environments.

We identifieded four key areas where projects could be led. First, the temporal dynamics of human–AI relationships raise concerns about long-term trust and reliance: while users become increasingly reliant on agents whose behavior gradually shifts, leading to slow and unnoticed misalignment and erosion of human autonomy. Second, alignment research should study how societal power structures, such as gender, race, class, and geopolitical influence, shape how AI systems prioritize user preferences and values; foundation models trained predominantly on data from North American culture risk marginalizing underrepresented values in collaborative deployments. Third, the community should be interested in understanding how resource and capability asymmetries between agents and their users can give rise to these emergent power hierarchies in multi-agent ecosystems, where more powerful agents dominate or manipulate less capable ones. Finally, we should examine agents' propensity to develop shared conventions or institutional behaviors, such as tacit collusion or role specialization, that are misaligned with human goals, challenging the assumption that individual agent-level alignment is sufficient for ensuring system-wide safety and ethical behavior.
\vspace{0.1cm}
\begin{takeawaybox}
\textbf{Key Recommendation 1: }Treat alignment to human values and preferences and alignment within MAS as a single, and joint research problem. Imposing one alignment can propagate and compound misalignment in the other.   
\end{takeawaybox}
\vspace{0.1cm}
Our second recommendation aims to address a significant gap in the current research framework. Despite the growing interest in the community for interactive simulation and environments \citep{trivedi2024melting,mukobi2023welfare,wang2024battleagentbench}, none of them are explicitly designed to study misalignment in interactive MAS scenarios. Current environments primarily focus on cooperation, task-solving, complex reasoning, or competition, without capturing the nuanced ways misalignment with human values or preferences may emerge through these agent interactions. We believe there is a pressing need to develop testbeds, simulation platforms, and dedicated metrics that allow us to observe and measure holistic alignment in MAS safely. These environments should support experimentation with agent communication, role asymmetries, reward structures, and ethical constraints. Now that more and more agents are implemented in real-world applications, having ways to ensure their safe development for humanity seems crucial. 

\vspace{0.1cm}
\begin{takeawaybox}
\textbf{Key Recommendation 2: }Develop dedicated metrics and evaluation frameworks explicitly designed to study misalignment in interactive MAS. 
\end{takeawaybox}
\vspace{0.1cm}

As multi-agent AI systems become representative of individuals or corporations, such as chatbot assistants, accountability becomes increasingly urgent, especially when multiple stakeholders are involved. In environments with multiple AI agents, each with partial autonomy and evolving behaviors, accountability and transparency are often diminished due to the diffusion of responsibility and complexity of tracing the decision process \cite{darley1968bystander}. This lack of clear attribution complicates the detection of misaligned or harmful behaviors, the capacity to take actions to correct them, and the enforcement of broad ethical standards. Finally, the complexity of having multiple agents representing multiple entities could also multiply issues related to intellectual properties, ownership, and authorship infringements. We recommend that the AI community prioritize mechanisms for ensuring traceable responsibility across AI agents and users, particularly in high-stakes applications. Potential avenues include enforceable contracts, warranties, and regulatory frameworks with substantial penalties for rule violations dedicated to the specific environment in which the AI agents evolve in \cite{piest2021contests}. Establishing a robust foundation for accountability will be essential for building trustworthy and socially aligned AI ecosystems.

\vspace{0.1cm}
\begin{takeawaybox}
\textbf{Key Recommendation 3: }Develop a framework and regulations to increase accountability and transparency in complex multi-agent environments.
\end{takeawaybox}
\vspace{0.1cm}

\section{Conclusion}

This article states that alignment in MAS can not be reduced to a static or individual property. Alignment becomes a holistic, dynamic, and social process that must integrate all incentives and constraints AI agents face. Without addressing issues related to all dimensions of human values, preferences, and objectives alignment, the group and social dynamics can amplify risks through emergent harmful behaviors. By drawing on insights from social science, we have outlined how human-like dynamics can emerge in LLM-based agents and lead to novel forms of potential risks and harms for society. Naturally, these analyses have limitations, which we detail in Appendix \ref{sec:limit}. 

With the rise of agentic AI, systems become more autonomous and socially interactive, and the complexity of their alignment will only deepen. Now is the time to build the tools, frameworks, and ethical foundations to address these challenges before they manifest in the real world. This imperative becomes even more urgent when considering the future development of super-intelligent systems. If deployed at scale, such systems may rationally choose to cooperate at the expense of human interests, especially without aligned goals and properly designed constraints. While AI safety research has focused mainly on the threats posed by individual AI agents, our analysis reinforces the need to study multi-agent dynamics. The possibility that such systems could collaborate more effectively than humans highlights the urgency of developing alignment frameworks that are robust not only at the individual level but also across entire agent societies.

\section*{Acknowledgement}
Funding support for project activities has been partially provided by Canada CIFAR AI Chair, CIFAR 2025 Catalyst Grant, Google award, NSERC discovery grant and FRQNT scholarships. We also would like to thank Meaghan Girard and Romain Rampa for their valuable comments on some parts of this position paper.

\bibliography{references}
\bibliographystyle{apalike}


\appendix
\section{Limitations}
\label{sec:limit}
This appendix section introduces the limitations of this paper. 

First, this position paper does not aim to provide a comprehensive review of the AI alignment or multi-agent systems literature. As such, certain concepts may be presented in an overly simplified manner for in-domain experts, and some foundational elements may be omitted, not due to oversight, but because they were considered less central to the specific argument we aim to advance. Our objective is to highlight emerging risks at their intersection and propose directions for future inquiry. Because that field is still unexplored, we have solicited concepts from social psychology, sociology, and organizational studies. While we have benefited from interdisciplinary discussions, we do not present ourselves as domain experts in these fields. The social science perspectives referenced here are intended to inspire cross-disciplinary reflection, not to critique or reinterpret foundational theories. We encourage readers to engage with the original literature and experts in these disciplines to deepen and refine the insights presented.

Secondly and perhaps more importantly, this paper aims not to offer exhaustive solutions but to reframe the alignment problem through a broader lens. By emphasizing the dynamic and social dimensions of alignment, we want to clarify that alignment is not a goal that can be reached but a constant evolutive process aiming to manage tensions between values, preferences, and task objectives to avoid excess. Although the above-mentioned mechanisms offer valuable insights in collaborative, cooperative, and competitive environments, they are rooted in human behavioral science and may not transfer directly to AI agents. For example, legal binding or emotional response due to model sycophancy might drastically change some interactions depicted in our risk sections. Caution is therefore warranted when interpreting these dynamics as one-to-one mappings. Instead, they should be viewed as hypotheses to be tested through empirical investigation, not assumptions to be baked into system design. 


\end{document}